\pdfoutput=1

\documentclass[11pt]{article}

\usepackage[preprint]{acl}

\usepackage{times}
\usepackage{latexsym}
\usepackage{float}

\usepackage[T1]{fontenc}

\usepackage[utf8]{inputenc}

\usepackage{microtype}

\usepackage{inconsolata}

\usepackage{graphicx}
\usepackage{xspace,mfirstuc,tabulary}
\usepackage{times}
\usepackage{adjustbox}
\usepackage{comment}
\usepackage{latexsym}
\usepackage{booktabs}
\usepackage{multirow}
\usepackage{comment}
\usepackage{xspace}
\usepackage{color, colortbl}
\usepackage{xcolor}
\usepackage{enumitem}
\usepackage{hyperref}
\usepackage{caption}
\usepackage{subcaption}
\usepackage{bm}
\usepackage{hyperref}
\usepackage{amsmath}
\usepackage{listings}
\usepackage{cleveref}
\definecolor{Color}{gray}{0.9}

\title{Charting the Landscape: Reviewing Five Years of African NLP Research}
\title{Charting the Landscape of African NLP: Mapping Progress and Shaping the Road Ahead}

\author{
    Jesujoba O. Alabi\textsuperscript{1} \,
    Michael A. Hedderich\textsuperscript{2} \,
    David Ifeoluwa Adelani\textsuperscript{3} \, 
    Dietrich Klakow\textsuperscript{1} \\
    \textsuperscript{1}Saarland University, Saarland Informatics Campus \\
    \textsuperscript{2}LMU Munich and Munich Center for Machine Learning \\ %
    \textsuperscript{3}Mila - Quebec AI Institute, McGill University \& Canada CIFAR AI Chair~~~\\
    \small{{\tt  \{jalabi,dietrich.klakow\}@lsv.uni-saarland.de,}\quad {\tt  mhedderich@cis.lmu.de,}\quad {\tt david.adelani@mila.quebec}}
}

\begin{document}
\maketitle
\begin{abstract}
With over 2,000 languages and potentially millions of speakers, Africa represents one of the richest linguistic regions in the world. Yet, this diversity is scarcely reflected in state-of-the-art natural language processing (NLP) systems and large language models (LLMs), which predominantly support a narrow set of high-resource languages. This exclusion not only limits the reach and utility of modern NLP technologies but also risks widening the digital divide across linguistic communities. Nevertheless, NLP research on African languages is active and growing. In recent years, there has been a surge of interest in this area, driven by several factors—including the creation of multilingual language resources, the rise of community-led initiatives, and increased support through funding programs. In this survey, we analyze 884 research papers on NLP for African languages published over the past five years, offering a comprehensive overview of recent progress across core tasks. We identify key trends shaping the field and conclude by outlining promising directions to foster more inclusive and sustainable NLP research for African languages.\footnote{We release our data and code publicly at \url{https://github.com/uds-lsv/africanlp-survey}}

\end{abstract}

\section{Introduction}
The field of Natural Language Processing (NLP) has undergone significant advancements in recent years. However, languages primarily spoken on the African continent have often been left behind in these developments. Many major large language models—such as Llama3~\citep{dubey2024llama3herdmodels}, Gemma2~\citep{gemmateam2024gemma2improvingopen}, and Anthropic’s Claude Sonnet-3.5~\citep{Anthropic2024}, do not list any African languages among their officially supported languages, thereby excluding the native tongues of potentially millions of speakers.

Despite this exclusion, NLP research on African languages is far from dormant. In recent years, there has been a surge of interest in both spoken and written form with the number of publications quadrupling in just five years and spanning a wide range of topics—from natural language understanding and text-to-speech systems to research on ethics, bias, and fairness. This momentum is fueled, in part, by community-driven initiatives, the creation of large corpora, shared tasks, and the emergence of dedicated venues.

\begin{figure}[t]
\setlength{\belowcaptionskip}{-10pt}
    \centering   \includegraphics[scale=0.30]{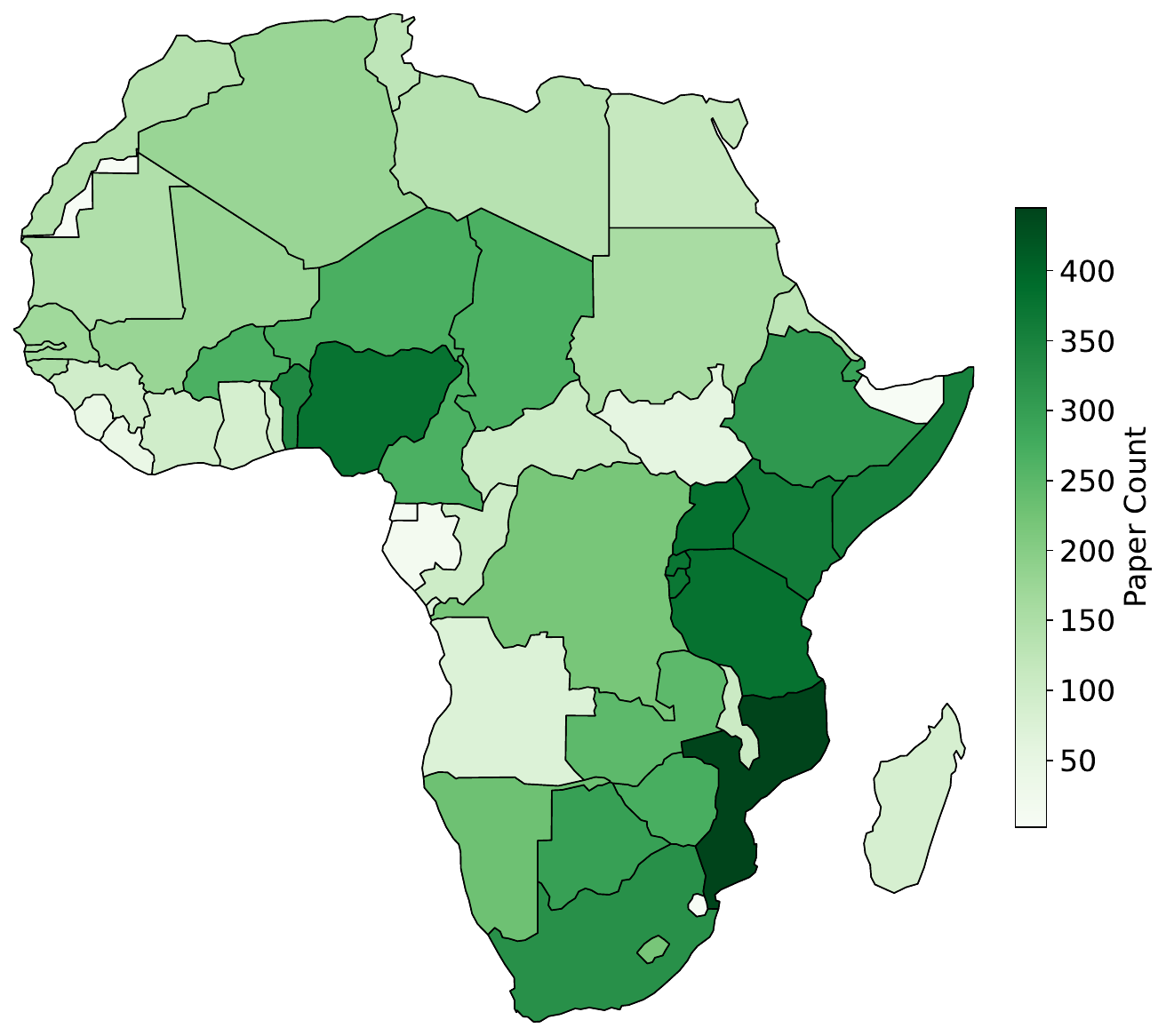}
    \caption{NLP research distribution across Africa by language coverage. Darker green indicate more papers on languages spoken in each country.\looseness-1}
    \label{fig:intro}
\end{figure}

While the broader NLP field has acknowledged the importance of supporting low-resource languages in general~\citep{bender2019BenderRule,joshi-etal-2020-state,hedderich-etal-2021-survey,nigatu-etal-2024-zenos}, this survey narrows its focus specifically to African languages. Many African countries share common historical, socio-economic, and infrastructural challenges, as well as pan-African research and policy initiatives. At the same time, Africa is characterized by its rich cultural and linguistic diversity~\citep{tchindjang2008languages,ouane2010and}. %
A continent-wide survey is therefore essential to capture both shared challenges and distinct differences in the development of language technologies for Africa.

\begin{figure*}[th]
\setlength{\belowcaptionskip}{-10pt}
    \centering   \includegraphics[scale=0.5]{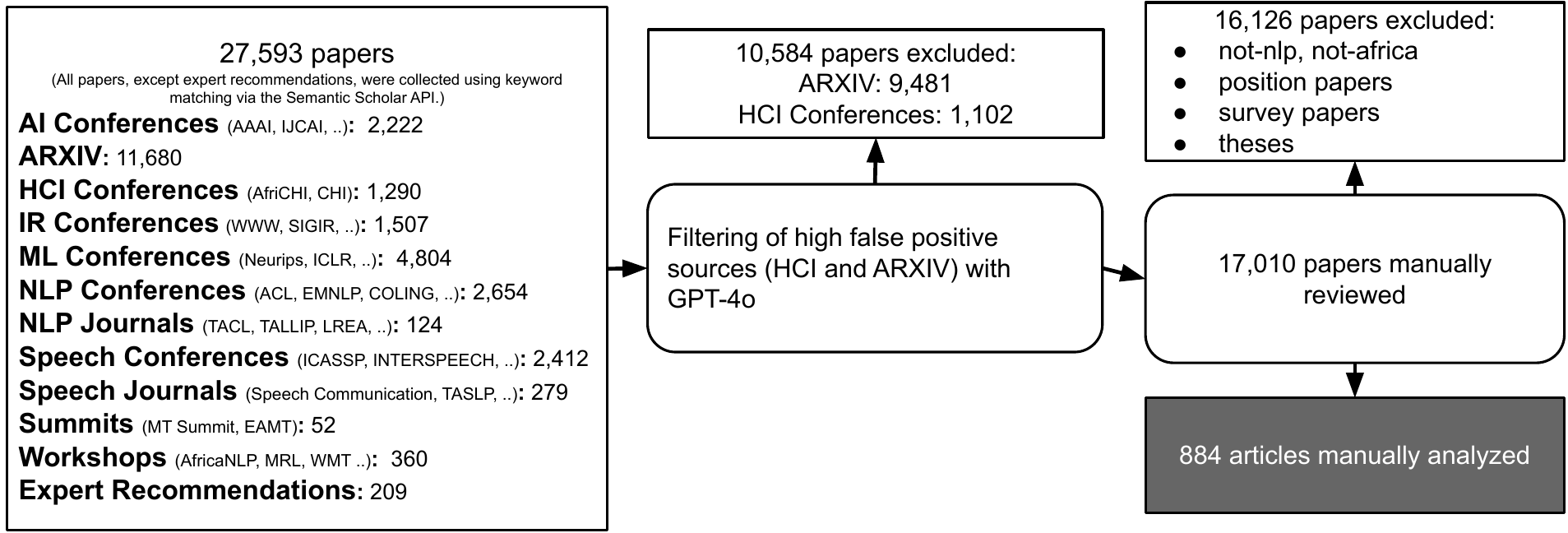}
    \caption{Inclusion flow of the systematic review.\looseness-1}
    \label{fig:pseudoflow}
\end{figure*}

This survey has three primary goals:

\begin{enumerate}
    \item \textbf{Comprehensive overview}: By systematically collecting and annotating literature from over twenty-five venues—spanning core NLP and speech, as well as adjacent fields like human-computer-interaction and machine learning— from the past five years, we provide an in-depth overview of contemporary research on African languages.
    \item \textbf{Accessible Entry Point}: By organizing this literature by language, task, technology, and theme, we offer a useful resource for both newcomers and experienced researchers newly engaging with African languages, thereby lowering the barrier to entry and encouraging further work in this crucial area.
    \item \textbf{Identification of Open Issues}: Through an extensive discussion of the current body of work, we highlight critical open issues—such as the imbalance in resources across African languages and the need for non-translated, native-language datasets—which can guide the strategic development of language technologies tailored to Africa.
\end{enumerate}

\section{Related Surveys}

African languages are usually classified as low-resource languages. In their six-class classification scheme, \citet{joshi-etal-2020-state} placed most African languages into the classes "left-behinds", scraping-bys", and "hopefuls" due to limited or non-existent labeled data. Surveys exist for low-resource languages in general, covering common NLP approaches to boost performance in settings with little data~\citep{hedderich-etal-2021-survey,haddow-etal-2022-survey,chen-etal-2023-empirical}. However, as \citet{nigatu-etal-2024-zenos} point out, the definition of "low-resource" is more complex than just a measurement of data availability, including also the availability of other resources and socio-political factors. This paper therefore focuses specifically on languages on the African continent, which could have substantially different settings compared to low-resource languages in other parts of the world.

Several studies have focused on languages within specific countries, such as South Africa~\citep{grover_2010}, Ethiopia~\citep{tonja-etal-2023-natural}, Kenya~\citep{amol2024statenlpkenyasurvey},  Nigeria~\citep{nwafor-andy-2022-survey,inuwadutse2025naijanlpsurveynigerianlowresource}, and Ghana~\citep{Azunre2021NLPFG,Issaka2024TheGN}. These surveys provide insights into the linguistic landscape, document available language resources, and explore computational methodologies tailored to the languages spoken in these countries. By offering a country-level perspective, they contribute to a broader understanding of language preservation and technological advancements within specific national contexts. Beyond country-specific studies, language-specific surveys have been conducted for individual African languages such as Southern Sotho~\citep{sibeko2022overview} and Yoruba \citep{jimoh2025bridginggapsnaturallanguage}. 
However, a survey looking at the broader African continent is necessary to uncover both unique and shared challenges of its languages, and chart a cohesive future for building technologies for African languages.

Similar to our effort, the survey by \citet{mussandi-wichert-2024-nlp} provides a tool-centric overview of NLP resources available for African languages. \citet{Keet2022BootstrappingNT} instead reviews how bootstrapping has been applied in practice to the Niger-Congo (‘Bantu’) subset of African languages, showing that rule- and grammar-based approaches transfer more successfully across distant languages, whereas data-driven methods work reliably only for closely related languages. In contrast, \citet{Ikae2024CurrentSO} concentrates on bias detection in machine translation for African and European languages. Our work differs by offering a broader landscape analysis of African NLP research, highlighting trends in language coverage, tasks, and research output.

\begin{figure*}[!t]
    \centering
    \begin{minipage}{0.32\textwidth}
        \centering
        \includegraphics[width=\linewidth]{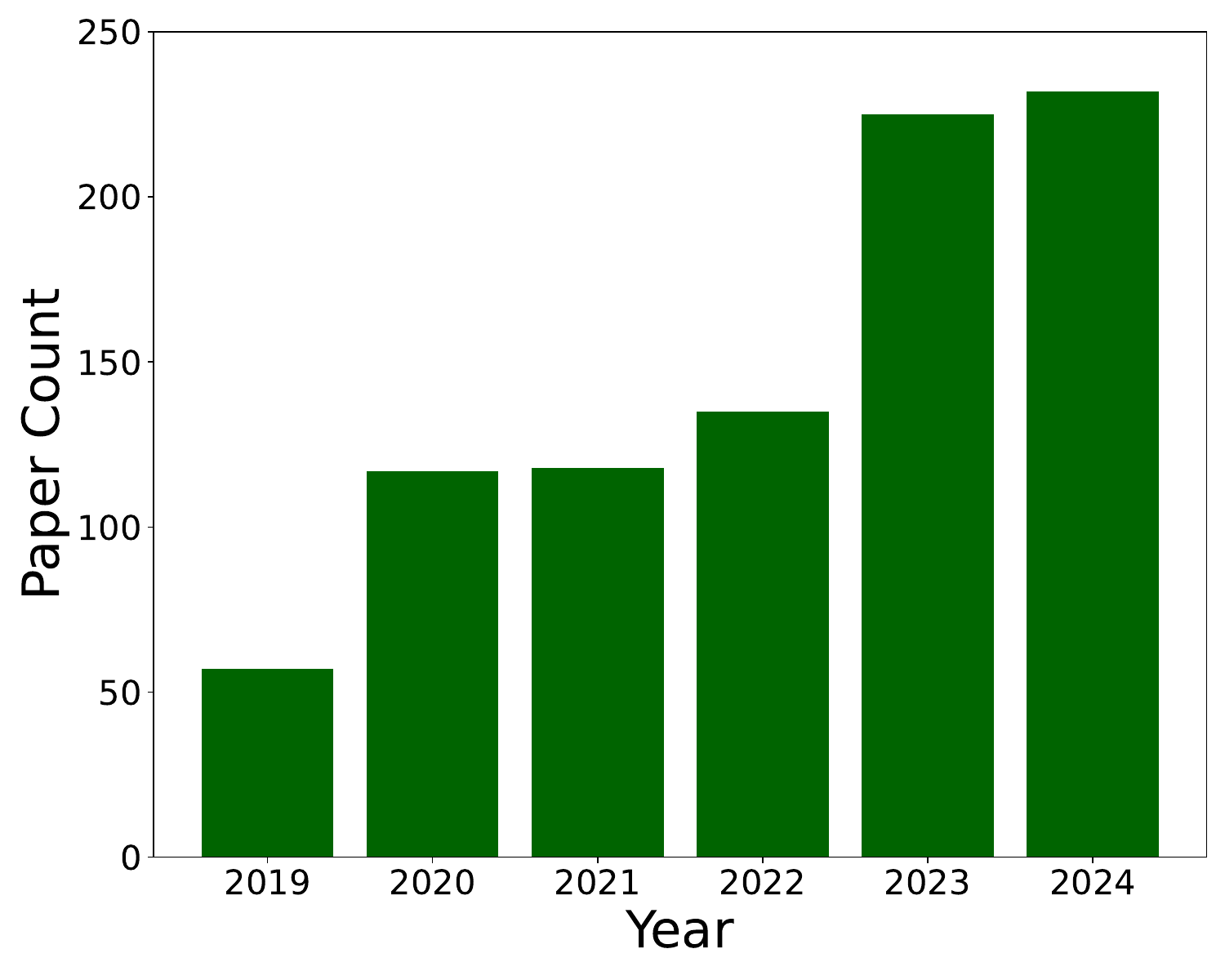}
        \caption{Distribution of papers by publication year.}
        \label{fig:paperdist}
    \end{minipage}
    \hfill
    \begin{minipage}{0.32\textwidth}
        \centering
        \includegraphics[width=\linewidth]{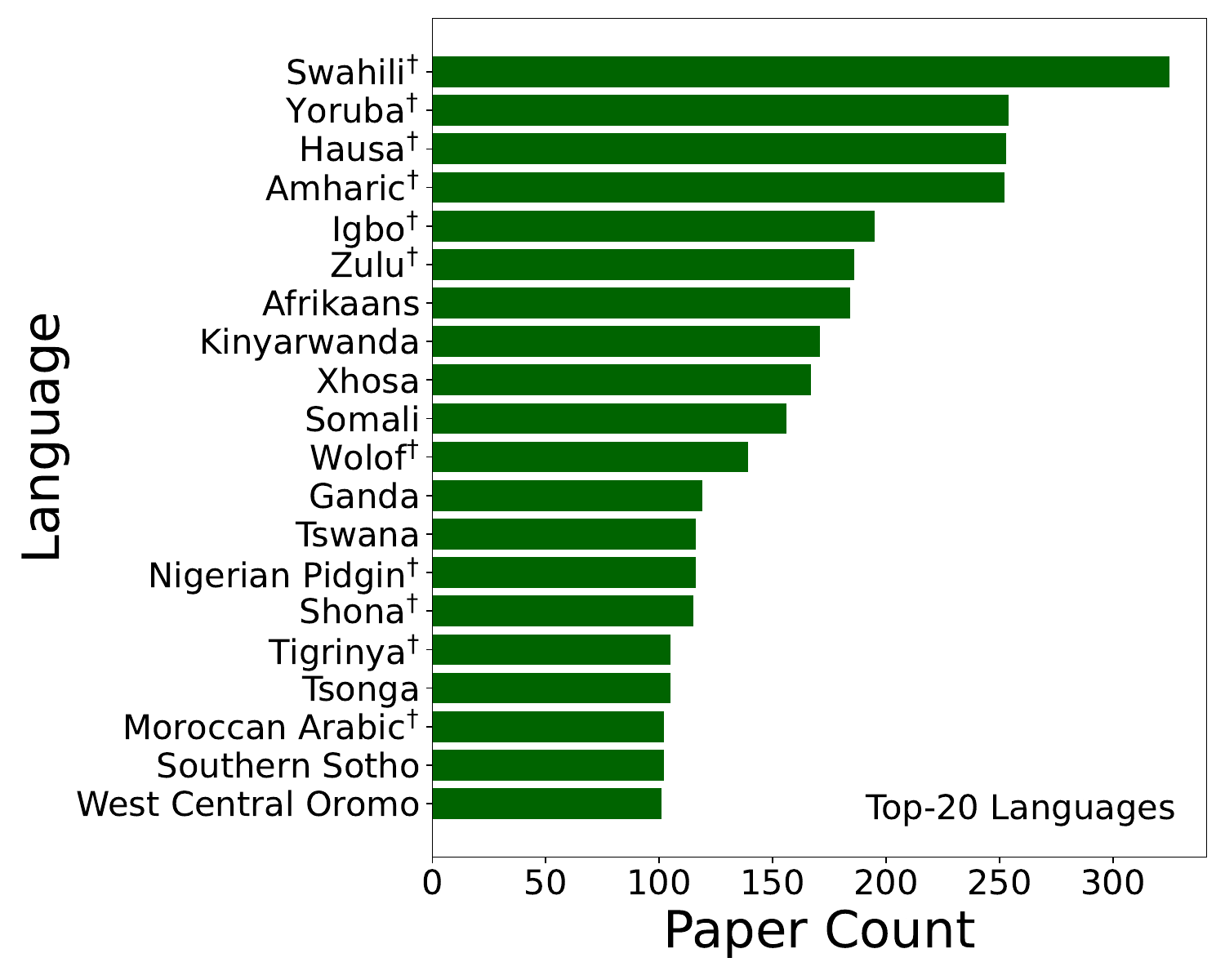}
        \caption{Top 20 languages by paper count;$\dagger$ indicates languages among the top 20 spoken in Africa.}
        \label{fig:toplang}
    \end{minipage}
    \hfill
    \begin{minipage}{0.32\textwidth}
        \centering
        \includegraphics[width=\linewidth]{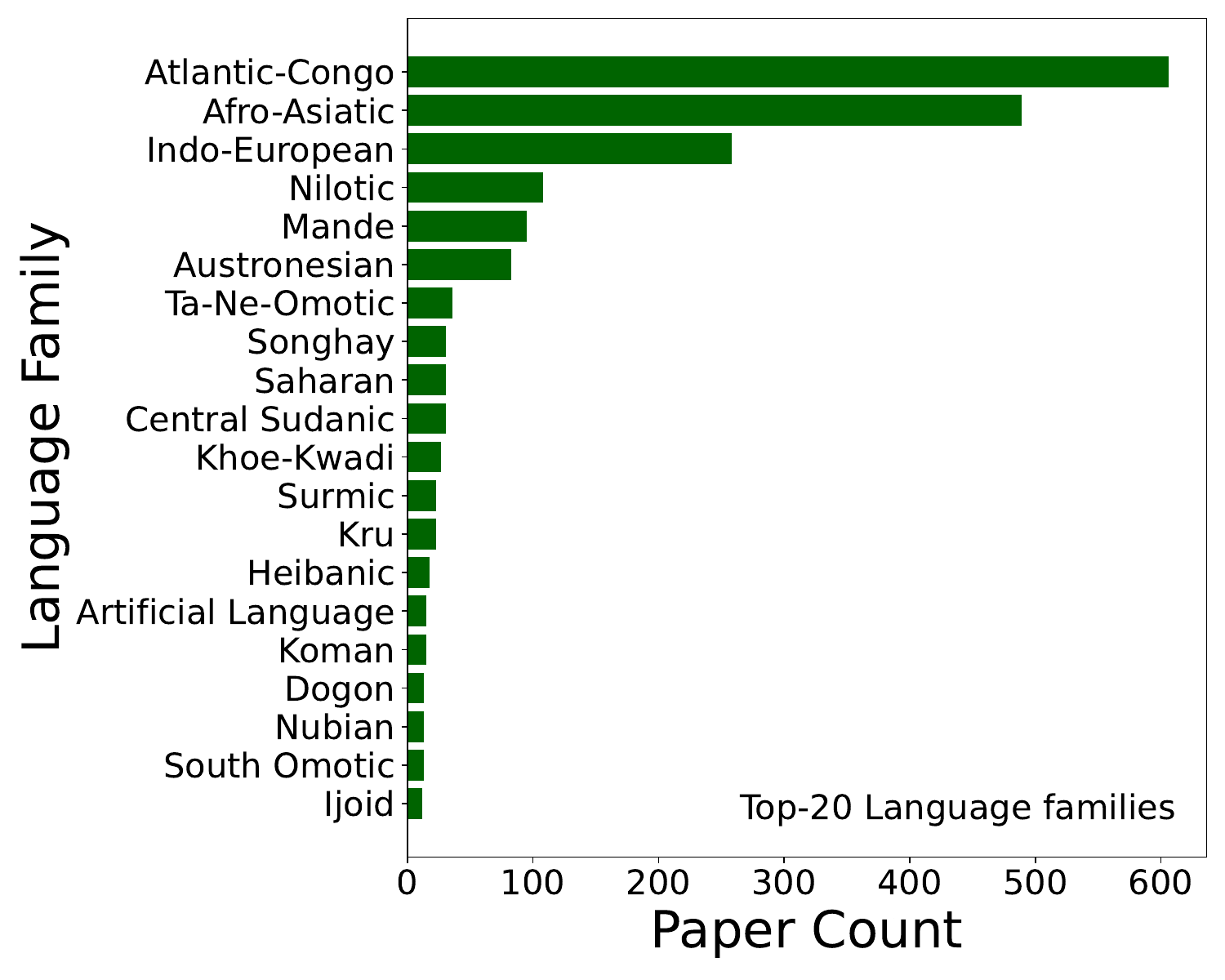}
        \caption{Top 20 language families by number of unique African languages studied.}
        \label{fig:topfamily}
    \end{minipage}
\end{figure*}

\section{Survey Methodology}
\paragraph{Literature Collection}
To conduct a comprehensive systematic survey of the relevant literature, we employ a multi-source approach that integrates automated and expert-driven methods. First, we use the \texttt{Semantic Scholar API}, which retrieves papers by matching query terms against titles and abstracts in its database. The inclusion criteria are “Africa”, the names of the 54 African countries, and 2,290 African languages from the Glottolog database.\!\footnote{Glottolog documents over 3,000 African languages and dialects, but we used only those with an official ISO 639-3 code.} 
These searches are conducted across leading conferences, journals, and workshops in NLP, speech, HCI, and AI, as well as the preprint repository arXiv. The full list of venues is provided in \Cref{app:venues}. We then augment this dataset with additional relevant papers to ensure the inclusion of influential and high-impact works that the \texttt{Semantic Scholar API} does not capture, adding 209 publications. In total, this process yields 27,593 relevant publications spanning five years, from 2019 to 2024.
By combining these sources, our approach ensures a well-rounded literature survey that balances algorithmic discovery with human expertise, leading to a more comprehensive and accurate understanding of the topic.

\paragraph{Filtering}
The automatically retrieved articles contained irrelevant articles not focused on AI or more specifically, NLP research, especially from HCI venues and pre-prints. To address this, we filtered the papers by prompting GPT-4o to classify whether they were NLP-related based on their title and abstract. Details for reproducibility are provided in~\Cref{app:datafilter}.

\paragraph{Manual Annotation}
We manually coded all the extracted papers with a codebook inspired by the ACL 2025 tracks and iteratively refined during the coding process. We identify the languages covered, the NLP tasks addressed, the low-resource techniques employed, and other relevant themes. In addition, our annotation process included determining whether the paper introduced a dataset and, if so, whether it was released and whether it was cross-cultural or translated from one language to another. We also assessed whether the paper proposed a model and, if so, whether the model was released.  We excluded papers on African American Vernacular English (AAVE) and those on spoken English that is not African-accent. We include African-accented variants of European languages because they are part of the linguistic reality of the continent, and handling them well is crucial for making NLP inclusive for Africans. \Cref{fig:pseudoflow} shows an overview of the data filtering process. The resulting publication set contains a total of 884 papers.

\section{Analysis Result}
In this section we analyze the annotated pool of publications along different dimensions.

\subsection{How much do researchers publish?}
\Cref{fig:paperdist} illustrates the distribution of the papers over the five-year period under review. The number of publications grew significantly—from approximately 57 papers in 2019 to 117 in 2020—continuing to rise and peaking at 232 papers in 2024. 
We hypothesize that this overall growth is driven by three key factors: (1) increased efforts to develop multilingual language resources, including embeddings and language models; (2) the emergence of community-led initiatives such as MasakhaneNLP,\footnote{\url{https://www.masakhane.io/}}  EthioNLP,\footnote{\url{https://ethionlp.github.io/}} HausaNLP,\footnote{\url{https://hausanlp.github.io/}} and,  GhanaNLP;\footnote{\url{https://ghananlp.org/}} and (3) the support of targeted funding programs such as Lacuna Fund\footnote{\url{https://lacunafund.org/}} and AI4D\footnote{\url{https://www.ai4d.ai/}}, which have provided crucial resources for African language research. Furthermore, shared tasks have contributed to this momentum, with one paper each in 2019 and 2020, 11 in 2021, 15 in 2022, 41 in 2023, and 24 in 2024. This upward trend highlights the growing and sustained interest in African NLP research.

\subsection{What languages do researchers work on?}
Given the annotated data, we evaluate all the African languages covered by the publications, using the African languages listed in \texttt{Glottolog}\footnote{\url{https://github.com/glottolog/glottolog-cldf/blob/master/cldf/languages.csv}}~\citep{hammarstrom2024bank} as a reference. Our analysis shows that a total of 2,275 African languages and dialects—including dialects of Arabic, creoles, and sign languages from various African countries—have been covered by the papers. However, only about 25\% of these languages appear in at least 10 papers, while over 67\% were featured in less than 5 papers, highlighting the skewed distribution of language coverage. \Cref{fig:toplang} presents the top 20 languages by frequency, which closely align with the rankings of the most spoken languages in Africa according to LinguaMeta~\citep{ritchie-etal-2024-linguameta-unified}. In particular, 11 of the top 20 languages according to publication counts appear among the most widely spoken languages. However, \texttt{Afrikaans} (7.3M speakers) and other South African languages such as \texttt{Tswana} (6M), %
and \texttt{Tsonga} (2.5M) appeared among the top 20 languages covered in the papers—demonstrating the significant research efforts dedicated to \texttt{South African} languages despite their relatively lower positions in the LinguaMeta ranking.
In terms of language families, \Cref{fig:topfamily} shows that \texttt{Atlantic–Congo} languages dominate the papers, followed closely by \texttt{Afro-Asiatic}, with other families receiving much less coverage.

In addition to indigenous African languages, some papers, particularly those focusing on speech also included African accents of widely spoken non-indigenous languages such as French, English, and Portuguese. While some studies clearly specified the country-specific accents they addressed~\citep{aryal2023sentimentanalysismultipleafrican,Afonja2021LearningNA,olatunji-etal-2023-afrispeech,hagemeijer-etal-2022-palma}, others were more generic~\citep{shan-etal-2023-english}. There are also research efforts on texts in non-indigenous languages tailored to African contexts, such as AfriSenti~\citep{muhammad-etal-2023-afrisenti} and AfriMed-QA~\citep{nimo-etal-2025-afrimed}.

\begin{figure}[t]
\setlength{\abovecaptionskip}{0pt}
\setlength{\belowcaptionskip}{-10pt}
    \centering   \includegraphics[scale=0.32]{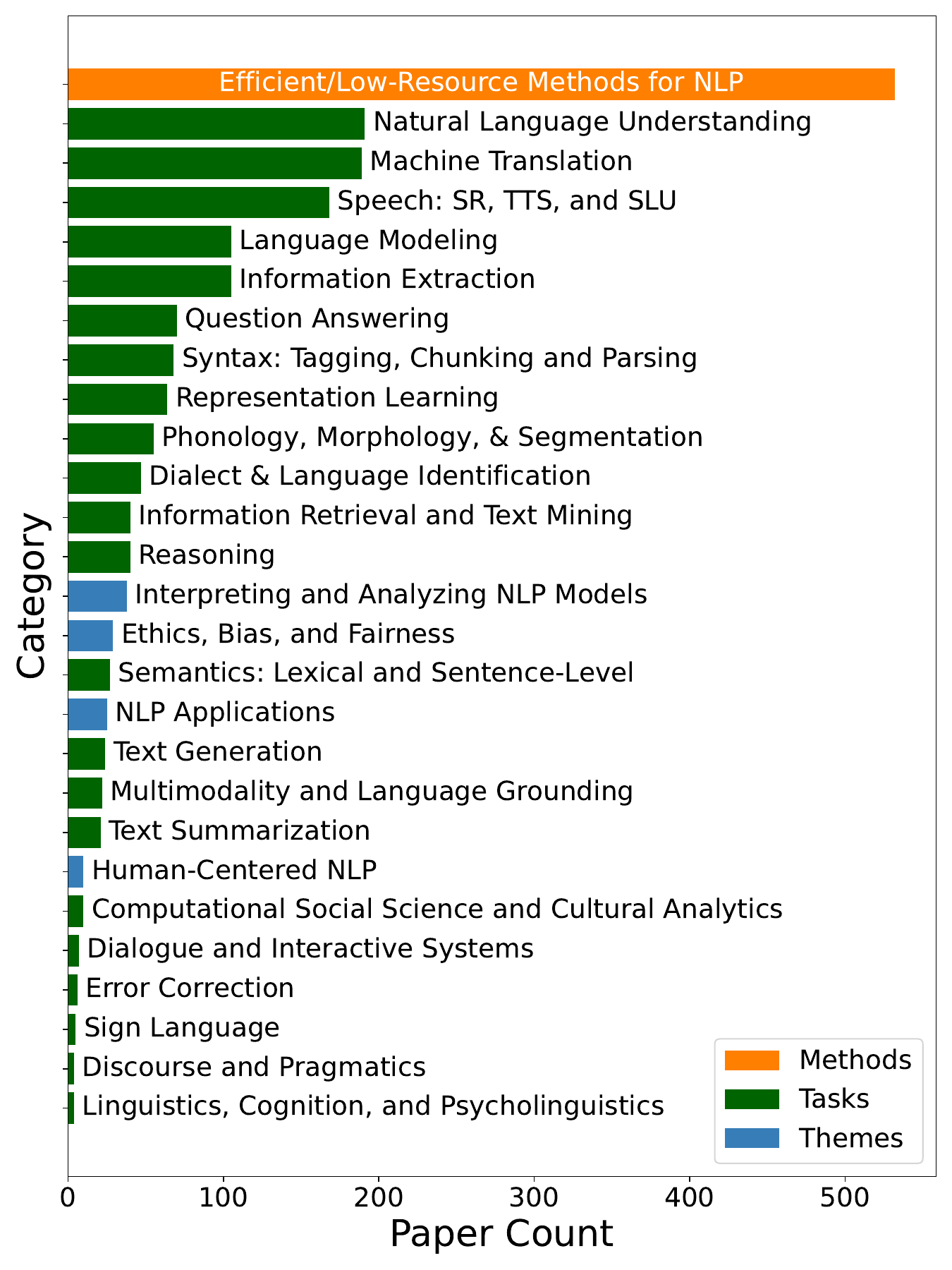}
    \caption{Bar plot of annotated paper categories (papers may appear in multiple categories).\looseness-1}
    \label{fig:taskplot}
\end{figure}

Furthermore, African languages are often code-mixed~\citep{bandia1996code}, typically with English or other languages spoken within the same region. Our analysis shows that 5.1\% of the reviewed articles focused on code mixing, with emphasis on the dialects of the North African Arabic dialects~\citep{hamed-etal-2023-data,hamed-etal-2024-zaebuc}, and South African languages~\citep{wilkinson-etal-2020-semi}, but also several other cases~\citep{Mountaga2021Bambara,muhammad-etal-2022-naijasenti,ilevbare-etal-2024-ekohate}.

Overall, in terms of continent coverage,~\Cref{fig:intro} shows that every country has at least one language covered, with Mozambique and Uganda leading at 445 and 383 papers respectively. Tanzania, Nigeria, Rwanda, Burundi, Kenya and Somalia also have strong representation, each with over 350 papers. This shows a notable skew in the geographic distribution of NLP research, with certain countries receiving more attention. 

\subsection{What tasks do researchers work on?}

The papers cover a skewed distribution of NLP tasks, with many addressing multiple tasks rather than just one. For instance, a paper that introduces a language model and evaluates it on a named entity recognition task contributes to both language modeling and named entity recognition. \Cref{fig:taskplot} shows the distribution of these tasks, and we highlight the 10 major tasks by publications below:

\textbf{Natural Language Understanding (NLU)} is the most common task category. Within this category, multiple subcategories of sequence classification tasks exist, such as text or topic classification~\citep{adelani-etal-2023-masakhanews,ma2023taxi1500}, hateful or offensive or abusive speech detection~\citep{Thirion2020TheSA}, emotion detection~\citep{moudjari-etal-2020-algerian,martin-etal-2022-swahbert}, natural language inference~\citep{lin-etal-2022-shot,ahuja-etal-2023-mega,adelani2025irokobenchnewbenchmarkafrican}, intent detection~\citep{moghe-etal-2023-multi3nlu,Skiredj2024DarijaBankingAN}, slot filling~\citep{fitzgerald-etal-2023-massive,Mastel2023NaturalLU}, semantic parsing~\citep{ruder-etal-2023-xtreme}, and sentiment analysis~\citep{Mountaga2021Bambara,muhammad-etal-2022-naijasenti,shode-etal-2023-nollysenti}. Notably, around 40\% of the NLU papers focus on sentiment analysis. Sequence classification tasks appear to be more common, likely due to the relative ease of creating datasets for them. This is especially true with techniques like annotation projection~\citep{adelani-etal-2023-masakhanews,ma2023taxi1500}, which facilitate multilingual data creation when parallel corpora are available. While some works focus on building datasets, others concentrate on developing methods to address these tasks, often through shared task contributions.

\begin{table}[t]
\small\centering
\scalebox{0.75}{
  \begin{tabular}{p{1.8cm}p{6.85cm}}
  \toprule
  \textbf{Task} & \textbf{Datasets} \\
  \midrule
  \multicolumn{2}{l}{\emph{Datasets}} \\
  NLU & Taxi-1500, SIB-200, MasakhaNEWS, AfriSenti, AfriXNLI \\
  MT & FLORES-200, AfroLingu-MT, NTREX-128, TICO-19, MAFAND-MT  \\
  Reasoning & AfriMGSM, LINGOLY, MGSM  \\
  IE & MasakhaNER, MasakhaNER 2.0 \\
  Speech & FLEURS, BibleTTS, African Voices  \\
  LM & mC4, ROOTS, WURA, GlotCC, MADLAD \\
  \midrule
  \multicolumn{2}{l}{\textit{Models}} \\
  MT & NLLB-200, Toucan, M2M-100, MADLAD \\
  Encoder LM & Serengeti, AfroXLMR\{-76L\}, AfroLM, AfriBERTaV2 \\
  Decoder LM & BLOOM, XGLM, mGPT, AfroLLaMa, InkubaLM \\
  Enc-Dec LM & Cheetah, AfriTeVa V2, AfriMBART, Afri\{M,By\}T5 \\
  \bottomrule
  \end{tabular}
  }
  \vspace{-2mm}
  \caption{Examples of 5 representative datasets and models with task categories, sorted by languages covered. 
  }
  \vspace{-5mm}
  \label{tab:works}
\end{table}

\textbf{Machine Translation (MT)}, the second most common task category, includes work on bitext mining and alignment~\citep{heffernan-etal-2022-bitext,feng-etal-2022-language}—especially from web sources—which has enabled the creation of widely used datasets like JW300~\citep{agic-vulic-2019-jw300}, CCAligned~\citep{el-kishky-etal-2020-ccaligned}, and WikiMatrix~\citep{schwenk-etal-2021-wikimatrix}, despite some concerns about content quality~\citep{kreutzer-etal-2022-quality}. Multilingual benchmarks like NTREX-128~\citep{federmann-etal-2022-ntrex}, Flores~\citep{goyal-etal-2022-flores,nllb2024scaling}, MAFAND-MT~\citep{adelani-etal-2022-thousand}, and AfroLingu-MT~\citep{elmadany-etal-2024-toucan} have also incorporated African languages. Furthermore, several efforts target specific language pairs~\citep{sanchez-martinez-etal-2020-english,nguer-etal-2020-sencorpus,adelani-etal-2021-effect,Azunre2021EnglishTwiPC}. Much of the work focuses on training and evaluating MT systems, including LLMs, for translation between African and high-resource languages~\citep{adelani-etal-2022-thousand,elmadany-etal-2024-toucan}. Furthermore, standard MT metrics like BLEU are inadequate for these morphologically rich languages, leading to the development of embedding-based metrics like AfriCOMET~\citep{wang-etal-2024-afrimte,wang-etal-2024-evaluating}, though its coverage remains limited.

\textbf{Speech Processing}: Aside from text-based research, there is a substantial amount of work focused on \textit{speech modality}, including automatic phoneme or speech recognition (ASR)~\citep{Dossou2021OkwuGbES,Mohamud2021FastDO,Conneau2022FLEURSFL,sikasote-etal-2023-big,ogunremi-etal-2024-iroyinspeech}, text-to-speech (TTS)~\citep{ogayo22_interspeech,meyer22c_interspeech,ma24c_interspeech,lux24_interspeech}, speaker recognition~\citep{villalba19_interspeech}, tone recognition~\citep{Obiang2024ImprovingTR}, emotion recognition, speech translation~\citep{zanon-boito-etal-2022-speech,sikasote-etal-2023-big,ahia-etal-2024-voices}, and acoustic modeling~\citep{Vyas2020LatticeFreeMA,Tachbelie2020DevelopmentOM}.
Recently, there has been growing interest in speech representation learning, particularly with transformer-based models. A few models include African languages, both African-centric~\citep{Kimanuka2023SpeechRD,caubriere2024ssaspeechssl,alabi25_interspeech} and massively multilingual models~\citep{Conneau2020UnsupervisedCR,boito2024mhubert,chen-etal-2024-towards-robust}. 
These advances are driven by efforts in the massive curation of speech resources.  However, despite the broad coverage of these models, the most commonly used evaluation benchmark for ASR, FLEURS~\citep{Conneau2022FLEURSFL}, covers only 21 African languages, creating an imbalance in evaluation.

\textbf{Language Models (LMs)}: During the survey period, there was a notable increase in the development of large transformer-based language models. Many studies introduced such models for African languages, including monolingual models for individual languages~\citep{nzeyimana-niyongabo-rubungo-2022-kinyabert,martin-etal-2022-swahbert}, and massively multilingual models that include African languages alongside others~\citep{Conneau2019,imanigooghari-etal-2023-glot500}. 
These models were developed by training them from scratch~\citep{jude-ogundepo-etal-2022-afriteva,ogueji-etal-2021-small} or extending existing language models~\citep{alabi-etal-2022-adapting,adelani-etal-2024-sib,meyer-etal-2024-nglueni}. These advances, like those in speech, rely on large-scale language data from the web. In this line are also works on representation learning, which includes learning embedding representation for text~\citep{yuan-etal-2020-interactive,liu-etal-2023-crosslingual-transfer} or speech~\citep{Jacobs2024MultilingualAW}, and analyzing them~\citep{alabi-etal-2020-massive}. In \Cref{sec:appmodels}, we provide an overview of several models for African languages, including LMs.

\textbf{Information Extraction (IE)} is another common category of NLP tasks and includes work on named entity recognition (NER)~\citep{adelani-etal-2021-masakhaner,adelani-etal-2022-masakhaner,rahimi-etal-2019-massively,Oyewusi2021NaijaNERC,Mbuvha2023MphayaNERNE}, entity linking and typing~\citep{lin-etal-2019-choosing,zhu-etal-2019-importance}, event extraction~\citep{jenkins-etal-2023-massively}, relation classification~\citep{lent-etal-2024-creoleval}, and keyword spotting and localization~\citep{Yusuf2019AnEE,Olaleye2022YFACCAY,Nortje2024ImprovedVP}. Among these, NER is particularly common due to the availability of resources such as MasakhaNER~\citep{adelani-etal-2021-masakhaner,adelani-etal-2022-masakhaner} and Wikiann NER~\citep{pan-etal-2017-cross}.

\textbf{Language and Dialect Identification}, for both text and speech, is another common task explored across many languages—including African languages—and has led to the development of tools such as AfroLID~\citep{adebara-etal-2022-afrolid}, GlotLID~\citep{kargaran-etal-2023-glotlid}, and OpenLID~\citep{burchell-etal-2023-open}. While some studies treat this as a benchmark task for evaluating representation learning models~\citep{boito2024mhubert,chen-etal-2024-fumbling}, others focus on developing practical tools~\citep{burchell-etal-2023-open,kargaran-etal-2024-glotscript-resource} that support downstream applications—such as multilingual data curation, where they are used to filter and organize resources by language~\citep{abadji-etal-2022-towards}.

\textbf{Question Answering (QA)}—which involves extracting, retrieving, or generating answers to input queries—remains an emerging research area for African languages. Recent efforts have focused on creating benchmarks from sources such as Wikipedia, for both QA and machine reading comprehension across multiple languages. For example, multilingual datasets like TyDi QA~\citep{clark-etal-2020-tydi}, which includes only Swahili from Africa, AfriQA~\citep{ogundepo-etal-2023-cross}, NaijaRC~\citep{aremu2024naijarc}, and AfriMed-QA~\citep{nimo-etal-2025-afrimed}. There have also been language-specific initiatives~\citep{Wanjawa2023kensquad,taffa-etal-2024-low,costa-jussa-etal-2025-y}. Another relevant work is Belebele~\citep{bandarkar-etal-2024-belebele}, a machine reading comprehension dataset translated into over 200 languages, including 50+ African languages. %

\textbf{Reasoning}: Closely related to QA is the growing focus on evaluating the reasoning abilities of large language models. Recent work explores various types of reasoning—linguistic~\citep{beanLINGOLYBenchmarkOlympiadLevel2024}, cultural knowledge~\citep{myung2024blend}, commonsense~\citep{ponti-etal-2020-xcopa}, mathematical~\citep{adelani2025irokobenchnewbenchmarkafrican}, comparative~\citep{agrawal-etal-2024-evaluating}, ethical~\citep{agarwal-etal-2024-ethical}, and moral~\citep{khandelwal-etal-2024-moral}. Notably, many of the datasets used for such evaluations in African languages are translated from English, which may limit cultural relevance and linguistic nuance. Evaluation results show that LLMs generally perform better on these tasks in high-resource languages compared to low-resource ones, including African languages~\citep{adelani2025irokobenchnewbenchmarkafrican}.

\textbf{Information Retrieval (IR)}--the task of finding relevant documents or information in response to a query—is a less common but growing area of research for African languages. Recent efforts include the development of multilingual and cross-lingual IR resources, such as Mr. TyDi~\citep{zhang-etal-2021-mr}, MIRACL~\citep{zhang-etal-2023-miracl} and CIRAL~\citep{adeyemi2024ciral}, which include one, two and four African languages respectively. Additional work in this field involves the creation of annotated resources for individual languages~\citep{Yeshambel2021MorphologicallyAA}. Furthermore, there are works on developing multilingual bitext and sentence retrieval models~\citep{artetxe-schwenk-2019-massively,feng-etal-2022-language,winata-etal-2024-miners} including several African languages.

\textbf{Syntax, Tagging, and Parsing}: A total of 68 papers have explored various syntax-related tasks, including dependency parsing~\citep{seddah-etal-2020-building,dione-2021-multilingual,steimel-etal-2023-towards,ralethe-2020-adaptation}, syntactic parsing~\citep{momoh-2024-lateral}, part-of-speech (POS) tagging~\citep{dione-etal-2023-masakhapos,faisal-etal-2024-dialectbench}, and computational grammar~\citep{Hellan2020ACG}. These efforts span both resource creation and evaluation; however, they cover only a limited number of African languages.

\textbf{Others:} Other tasks categorization with less publications including semantics, multimodality (text + image or video), text generation (excluding machine translation) are described in \Cref{app:othertasks}. 

\subsection{What resources are available?}
Given that African languages are low-resource, efforts over the past five years have focused on developing resources—both for individual languages and at a large-scale multilingual level, particularly in the form of datasets. Our analysis shows that 401 papers (about 45\%) involve dataset creation, with 126 papers creating datasets via translation. Many large multilingual datasets—such as NLU’s SIB-200 and Taxi-1500—are based on English translations. Similarly, speech datasets like FLEURS, FLEURS-R, and SpeechTaxi rely on translated data, which often lacks cultural relevance and may exhibit translationese~\citep{koppel-ordan-2011-translationese,bizzoni-etal-2020-human}, limiting their real-world applicability. \Cref{tab:works} lists examples of datasets and models, with a more comprehensive list in \Cref{app:datalist}.

\subsection{What are the efficient or low-resource techniques used?}
Given the low-resource status of African languages, researchers often rely on specialized techniques. \Cref{fig:lowrestech} (\Cref{app:lowrestec}) summarizes those used in the analyzed papers; we highlight six below.

\textbf{Transfer Learning} Our analysis shows that about 49\% of the papers used transfer learning, primarily through word embeddings, or by fine-tuning pretrained models or using them as feature extractors. This approach has been effective across many languages, including low-resource ones, largely due to the availability of multilingually pretrained models that support cross-lingual transfer~\citep{pfeiffer-etal-2021-unks}. However, a major challenge arises when the target African language is not well represented in the pretrained model or its tokenizer. Common strategies to address this include \textbf{adaptive pretraining}~\citep{pfeiffer-etal-2021-unks,alabi-etal-2022-adapting,adelani-etal-2024-sib,meyer-etal-2024-nglueni}, and embedding initialization~\citep{liu-etal-2024-ofa,Quinjica2024ANGOFALO,dobler-de-melo-2023-focus}.

\textbf{Zero-Shot Cross-Lingual Transfer Learning} is another commonly used technique in the papers. In this approach, a model is trained—often using transfer learning—on a source language and then evaluated directly on the target language. While English is a commonly used source language, some of the analyzed papers explore how to select the most effective source language~\citep{dione-etal-2023-masakhapos,adelani-etal-2023-masakhanews} and how to improve performance from the model side such as checkpoint and run averaging~\citep{schmidt-etal-2023-free,schmidt-etal-2023-one}.

\textbf{Data Augmentation} Data augmentation is another popular low-resource technique, used in  8\% of the  analyzed papers, and it has been successfully used for various tasks such as language modeling~\citep{adelani-etal-2024-sib,singh-etal-2024-aya,azime-etal-2024-walia}, machine translation via back translation~\citep{reid-etal-2021-afromt,adelani-etal-2021-effect}, and other tasks~\citep{zhang-etal-2024-aadam}. %

\textbf{Audit and Data Filtering} Due to the presence of language resources on the web that often contain irrelevant or low-quality content, auditing and filtering have become common and effective approaches—especially for low-resource languages. For instance, these methods have been shown to improve performance in language modeling tasks~\citep{oladipo-etal-2023-better, Kudugunta2023MADLAD400AM}

\textbf{Weak and Distant Supervision} was also employed in the reviewed works, with the aim to improve model performance by generating noisy labels for unlabeled data via heuristics or external sources. When paired with noisy-label handling approaches, it proves beneficial for African languages.~\citep{Zhu2022TaskAdaptivePF,hedderich-etal-2020-transfer,Adelani2020DistantSA}.

\textbf{Other methods} Other techniques including annotation projection (AP), meta learning, and multi-task learning (MTL) are described in \Cref{app:lowrestec}.

\subsection{What are the other themes?}

\textbf{Interpretability and Analysis (IA)}:
Methods on interpretability and model analysis have gained increasing attention in NLP in recent years~\citep{mosbach-etal-2024-insights}. Although few studies focus exclusively on African languages~\citep{chimoto-etal-2024-critical}, there are several multilingual IA works on MT~\citep{Ahia-etal-2021-low-resource,Adebara-etal-2022-linguistically,Mohammadshahi-etal-2022-compressed}, speech models~\citep{Osakuade2024DoDS},  and LLM~\citep{ahia-etal-2023-languages,alabi-etal-2024-hidden,Zebaze2024InContextES,zhang-etal-2024-impact,shafayat2024multifact} which include African languages. We identified 38 relevant papers in total, showing progress but highlighting the need for further work, especially in applying these insights to improve models' performance.

\textbf{Ethics, Bias, and Fairness}: There are also a few works that focus on analyzing biases in NLP models and proposing mitigation strategies to promote fairness. Among the 29 relevant papers identified,  gender bias~\citep{Costajuss2022OccGenSO} and cultural bias~\citep{shan-etal-2023-english,olatunji23_interspeech,Magomere2024TheWW} were the most commonly addressed. However, much remains to be done to address bias and ensure equity across diverse linguistic and cultural contexts.

\textbf{Human-Centered NLP}: Human-centric research in NLP has gained prominence in both the NLP community~\citep{hucllm-2024-human} and the field of human-computer interaction (HCI), which has seen a massive uptake on work about LLMs~\citep{pang25LLMificiation}. However, such work remains scarce in African NLP—for example, only 11 relevant papers have been presented at CHI, a leading HCI conference, and its African counterpart, AfriCHI.

\textbf{NLP Applications}: Our analysis identified 25 relevant papers that focus on the development of NLP technologies aimed at application areas such as education~\citep{Corallo2023OpticalCR}, health~\citep{Abdulhamid2023CanLL}, agriculture~\citep{Akera2019KeywordSM}, banking~\citep{Skiredj2024DarijaBankingAN}, and humanitarian response~\citep{ktem2021CongoleseSM}. %

\section{Discussion and Future Directions}
In this survey, we presented an overview of research on African languages over the past five years, covering key tasks such as machine translation, sentiment analysis, language identification, question answering, and reasoning with large language models. While notable progress has been made—particularly in dataset creation, benchmarking, and adaptation of multilingual models—our analysis reveals several limitations and points to important future directions.

\textbf{Scaling beyond the top-10 resourced languages}: It is important to invest efforts in supporting African languages beyond \textit{Swahili} and other widely spoken ones, ensuring that lower-resourced and endangered languages also receive attention in preservation, research, and technological integration. Unlike other regions, Africa has more than 100 languages with more than 1M speakers, so scaling beyond the top-10 is urgently needed. 

\textbf{Towards more multi-cultural dataset creation}: Our analysis shows that many benchmark datasets created for these languages, especially large-scale datasets such as \textit{Flores}, are based on translated texts.  Some of these lack cultural contexts of the users of this technology, and can amplify bias towards Western concepts especially for QA tasks~\citep{romero2024cvqa} or reduce performance due to lack of cultural understanding~\citep{akinade-etal-2023-varepsilon,yao-etal-2024-benchmarking}. There is a need to go beyond scaling by translation to building realistic multicultural datasets for African languages. 

\textbf{Towards more multimodal models for African languages}: At the moment, there is significantly less research on developing \textit{Visual LMs} (VLMs) for African languages. Existing VLMs perform poorly on African languages and cultural understanding of artifacts from African countries~\citep{nayak-etal-2024-benchmarking,Winata2024WorldCuisinesAM}. This makes it difficult to answer culture-specific questions over images, and also amplifies cultural bias related to Africans identity, lifestyles, occupations, food, clothing, among others~\citep{Chiu2024CulturalBenchAR}. %

\textbf{More investment in speech processing tasks}: Many languages in Africa are often  \textit{spoken} rather than \textit{written}. However, only few languages have large-scale datasets covering over 100 hours for ASR. While platforms such as Mozilla CommonVoice have democratized the data collection, only few languages used the platforms due to lack of attribution by the contributors of the datasets~\citep{Rajab2025TheEF} and the difficulty of finding volunteers. Aside from popular tasks like ASR, other tasks such as text-to-speech, audio classification, speech translation, and many others lack datasets. To develop NLP models that would be used by many African communities, it must be accompained with a speech component.

\textbf{More exploration of other generation tasks}: Most benchmarks for African languages have primarily focused on NLU and MT, partly due to the relative ease of creating datasets for these tasks. However, more attention should be given to language generation tasks—such as text summarization, grammar correction, dialog and table-to-text generation—which are typically more challenging to build datasets for but are equally important for the development of robust language technologies.

\textbf{More development of African-centric LLMs}: Current LLMs, especially the open-weight versions, often do not officially include African languages in their pre-training~\citep{Aryabumi2024Aya2O}, and when they do, only few high-resource ones such as Afrikaans and Swahili with more than 1B tokens are covered~\citep{,qwen3_paper}. There is need to develop or adapt LLMs to more African languages. This should be accompanied by evaluation datasets that are culturally relevant, sector-specific, knowledge-intensive, and require reasoning to assess LLMs in truly low-resource languages.

\textbf{Towards human-centered, application-driven NLP}: Understanding human-centric factors and developing methods in the context of applications is essential to ensure that NLP technologies genuinely benefit users. More research is needed to explore the shared and distinct human-centric needs of African language-speaking communities, both within the continent and in comparison to other global contexts.

Finally, the successful implementation of all the aforementioned directions ultimately depends on activities such as support for community initiatives, the organization of shared tasks, and adequate funding—all of which serve as the foundational pillars for initiating, sustaining, and scaling these efforts.

\section{Conclusion}
In this survey, we provide a structured overview of recent developments in natural language processing for African languages over the past five years. Our study reveals steady and significant progress in research on African languages, driven by several factors such as the creation of multilingual language resources, increased community engagement, and various funding initiatives. We highlight common tasks, methods, and recurring themes that characterize current research efforts. However, these efforts have been disproportionately focused on a limited set of languages, tasks, and countries. As a result, we outline several potential directions for future research to ensure more balanced and inclusive development across the continent.

\section*{Limitations}

\paragraph{Coverage Limitations:} Although we used a systematic survey approach that combines automated and expert-driven methods, one of the limitations of this survey is that it primarily focused on papers published in top-tier conferences and journals, which may exclude relevant works published within African venues and other venues. While we included papers from AfricaNLP, and AfriCHI, which are both African-centric venues, local research published in other venues may offer additional insights that are not captured in this study. Future work should broaden the scope to incorporate Africa-specific studies to provide more comprehensive understanding of the field and ensure that insights from the different parts of Africa are adequately represented. Similarly, our automated method, which relied on the Semantic Scholar API, may not have correctly indexed all relevant papers from the selected venues. 

\textbf{Potential False Negatives in Data Selection:} Our manual annotation of 100 examples revealed that our filtering process using GPT-4o on HCI and arXiv papers resulted in a false negative. %
It is possible that a few relevant NLP papers on African languages were also omitted, particularly since large language models like GPT-4o are sensitive to prompts, and we only tested a single prompt.

\textbf{Lack of dataset sheet for some papers:} Some papers such as ~\citep{li22aa_interspeech}, ~\citep{Nguyen2024MultilingualDI} and toolkit papers such as ~\citep{Emezue2020LanfricaAP} and \citep{Muite2021AnOS} did not provide the list of languages covered despite covering hundred and thousands of languages. Although we included them in our analysis, we could not provide the studied languages. 

\textbf{Lack of social and ethical dimensions:}  of African NLP development. This survey focuses primarily on the technical landscape of African NLP, including datasets, tasks, languages, and modeling approaches, and does not systematically address social, cultural, or ethical dimensions such as community involvement, consent, bias, or potential harms of NLP systems. Human-centered evaluations and assessments of societal impact are beyond the scope of this survey. Future research should incorporate ethical, cultural, and social considerations to ensure that African NLP technologies are developed responsibly and equitably.

\section*{Acknowledgments}
Jesujoba Alabi was funded by the Deutsche Forschungsgemeinschaft (DFG, German Research Foundation) – Project-ID 232722074 – SFB 1102. David Adelani acknowledges the funding of IVADO and the Canada First Research Excellence Fund.  We are grateful to Aravind Krishnan and Max Rausch-Dupont for their feedback on the manuscript. Finally, we would like to thank OpenAI for providing API
credits through their Researcher Access API Program. 

\bibliography{custom,latex/anthology}

\appendix

\section{Data collection venue}
\label{app:venues}
We collected data from 11 venue categories, as shown in \Cref{tab:data_stats}. While we report both the total number of papers initially retrieved and those ultimately analyzed, not all venues had retrievable papers, and not all contributed to the final analysis (for example, CSCW, UIST). The ACL Summit category had the fewest papers retrieved, while ArXiv, as expected, had the most. Due to the size, and presence of several irrelevant papers from ArXiv and HCI venues, we filtered them using GPT-4o, as described in \Cref{app:datafilter}. In addition to the automatically retrieved papers, we also included expert-recommended papers from various venues. Here, experts refers to African researchers who have contributed to African language research over the past five years.

The following are full names of the conferences, journals and workshop names
\begin{enumerate}
    \item AAAI: Association for the Advancement of Artificial Intelligence
    \item CHI: ACM Conference on Human Factors in Computing Systems
    \item ACL: Annual Meeting of the Association for Computational Linguistics
    \item EACL: European Chapter of the Association for Computational Linguistics
    \item LREC: International Conference on Language Resources and Evaluation
    \item EMNLP: Conference on Empirical Methods in Natural Language Processing
    \item COLING: International Conference on Computational Linguistics
    \item ELRA: Language Resources and Evaluation Journal
    \item TACL: Transactions of the Association for Computational Linguistics
    \item TALLIP: ACM Transactions on Asian and Low-Resource Language Information Processing
    \item RAIL: Workshop on Resources for African Indigenous Languages
    \item ICASSP: International Conference on Acoustics, Speech, and Signal Processing
    \item INTERSPEECH: International Speech Communication Association
    \item SLT: IEEE Spoken Language Technology Workshop
    \item AfricaNLP: Workshop on African Natural Language Processing
    \item DeepLo: Workshop on Deep Learning Approaches for Low-Resource NLP
    \item MRL: Workshop on Multilingual Representation Learning
    \item WOAH: Workshop on Online Abuse and Harms
    \item WMT: Conference on Machine Translation
    \item EAMT: European Association for Machine Translation
    \item MT Summit: Machine Translation Summit
    \item TASLP: IEEE/ACM Transactions on Audio, Speech, and Language Processing
    \item NeurIPS: Conference on Neural Information Processing Systems
    \item ICLR: International Conference on Learning Representations
    \item ICML: International Conference on Machine Learning
    \item CSCW: Conference on Computer-Supported Cooperative Work \& Social Computing
    \item UIST: The ACM Symposium on User Interface Software and Technology
    \item CIKM: The Conference on Information and Knowledge Management
    \item SIGIR: Special Interest Group on Information Retrieval
    \item TREC: Text Retrieval Conference
    \item WWW: International World Wide Web Conference
    \item FAccT: ACM Conference on Fairness, Accountability, and Transparency
    \item RANLP: International Conference Recent Advances in Natural Language Processing
\end{enumerate}
\begin{table*}[t]
\begin{center}
\footnotesize
\begin{adjustbox}{width=\textwidth,center}

\begin{tabular}{l|l|cc}
\toprule
\textbf{Category} & \textbf{Venues} & \textbf{\#papers} & \textbf{\#relevant} \\
\midrule
\multicolumn{4}{l}{\textit{Semantic Scholar}} \\
AI Conferences	& AAAI, IJCAI	& 2222 & 6 \\
ARXIV	& arXiv.org	& 11680 & 188 \\
HCI Conferences	& CHI, AfriCHI, CSCW, UIST	& 1290 & 8 \\
IR Knowledge	& CIKM, SIGIR, TREC, WWW	& 1507 & 8 \\
ML Conferences	& Neurips, ICML, ICLR, FAccT	& 4804 & 10 \\
NLP Conferences	& ACL, COLING, EACL, EMNLP, LREC, NAACL, RANLP	& 2654 & 209 \\
NLP Journals	& LREA, TACL, TALLIP	& 124 & 14 \\
Speech Conference	& ICASSP, INTERSPEECH, SLT	& 2412 & 92 \\
Speech Journal	& TASLP, Speech Communication	& 279 & 11 \\
Workshops	& AfricaNLP, DeepLo, MRL, WOAH, WMT	& 360 & 124 \\
ACL Summits	& EAMT, MT Summit	& 52	& 5 \\
\multicolumn{4}{l}{\textit{Expert Collection	}} \\		
	{} & Same venues as above, and others e.g. RAIL	& 209	& 209 \\
\midrule
Total	& {} &	27617 & 884  \\

\bottomrule
    \end{tabular}
\end{adjustbox}
\caption{\textbf{Data statistics} the venues considered for paper collection, the number of unique papers retrieved, and the number of papers that are relevant after annotation.}
\label{tab:data_stats}
  \end{center}
  \vspace{-2mm}
\end{table*}

\section{Data filtering}
\label{app:datafilter}
Due to the large volume of papers retrieved from ArXiv and the presence of many irrelevant ones, we opted for a coarse filtering approach rather than a fine-grained one. Specifically, we filtered papers based on whether they were related to natural language processing. To assist with this process, we used GPT-4o. Additionally, we included papers from HCI conferences in this filtering step. 
To address this, we filtered the papers by prompting GPT‑4o\footnote{the 2024-08-06 version} to classify them into two categories—relevant and not-relevant—based on their title and abstract (our prompt is provided in \Cref{lst:prompt}). We then manually evaluated 100 papers from the automatic filtering—50 from each category—and obtained an accuracy of 99\%, with all errors being false negatives.

\begin{figure*}[t]
\centering
\begin{lstlisting}[caption={Prompt used for filtering the ArXiv and HCI papers}, label={lst:prompt}]
You are an expert research assistant in Natural Language Processing (NLP).

Given a paper's title and abstract, decide whether the paper should be classified as "relevant" or "not relevant".

A paper should be classified as relevant if it is about:
    - Natural language processing (NLP) or computational linguistics
    - Speech processing (e.g., recognition, synthesis, spoken dialogue)
    - Multimodal tasks involving language (e.g., vision-language, speech-text, OCR with text processing, audio-text)
    - Dataset curation, annotation, or benchmarks for NLP/multimodal tasks
    - Core NLP tasks (e.g., text classification, parsing, entity linking, translation, summarization, information extraction, question answering, dialogue systems, sentiment analysis, semantics, discourse, topic classification, misinformation detection)
    - Information retrieval, text mining, or data mining involving text or language understanding
    - Evaluation metrics for NLP tasks (e.g., BLEU, ROUGE, COMET, etc.)
    - Language modeling
    - Language generation
    - Ethics, bias, fairness, interpretability, safety in language technology systems
    - Human-centered NLP (e.g., user-centered design of language technology, human-LLM interaction, social impact of language technology)
    - Applied NLP in specialized domains (e.g., agriculture, biomedical, legal, education, cultural analytics, computational social science, NLP for social good)

Otherwise, the paper should be classified as not relevant.

Output only one label: "relevant" or "not relevant".

Title: "{title}"

Abstract: "{abstract}"
 \end{lstlisting}
\end{figure*}

\section{Other tasks researchers work on}
\label{app:othertasks}

\textbf{Semantics} Work on semantics include semantic textual relatedness~\citep{ousidhoum-etal-2024-semeval} which was run as a shared task, lexical semantic relations~\citep{gromann-etal-2024-multilexbats}, and word analogy tasks~\citep{mersha-wu-2020-morphology,Gaustad2023ExploringAW} mostly in the context of word embeddings. 

\textbf{Phonology and Morphology} We observed from our analysis that there are papers under phonology, such as those on phonetics~\citep{issa23_interspeech,ahn23_interspeech} and syllabification~\citep{sibeko-setaka-2023-evaluating,Sibeko2024DevelopingAT}, studying these phenomena in both text and speech. In the area of morphology, some papers focus on lemmatization~\citep{Mohamed2023LexiconAR}, morphological segmentation and analysis~\citep{nzeyimana-2020-morphological,Puttkammer2021CanonicalSA}, and under word segmentation, on tasks like tokenization~\citep{Atuhurra2024IntroducingST,meyer-buys-2022-subword}. 

\textbf{Dialogue and Interactive Systems} We found a few papers in this category, including the creation of dialogue datasets for some African languages~\citep{Adewumi2023AfriWOZCF}, conversational data for dialects of English such as Nigerian English~\citep{Eisenstein2023MD3TM}, and the development of chatbots~\citep{Mabrouk2021AMA,Hailu2024DeepLB}.

\textbf{Text with Images or Video}: 
There are a few works on multimodality that combine text with images or video, such as image captioning~\citep{abdulmumin-etal-2022-hausa}, keyword localization~\citep{Olaleye2022YFACCAY,Nortje2024ImprovedVP}, multimodal machine translation (MT)~\citep{abdulmumin-etal-2022-hausa,hatami-etal-2024-english}, text-to-image generation~\citep{leong-etal-2022-bloom,Magomere2024TheWW}, and visual question answering~\citep{leong-etal-2022-bloom}. A notable contribution is the development of a multimodal LLM for Amharic~\citep{Andersland2024AmharicLA}. Sign languages, which are inherently multimodal due to their use of visual and spatial cues, have also received attention, with few resource-focused papers addressing sign language recognition~\citep{Elhagry2021EgyptianSL}, sign-to-speech~\citep{ijcai2022p855}, and sign language translation~\citep{Gueuwou2023AfriSignMT,gueuwou-etal-2023-jwsign}.

\textbf{Language Generation}: Our analysis shows that sequence generation tasks—such as text generation~\citep{Chang2020UnsupervisedPT,Ramalepe2023TheAO}, text summarization~\citep{ouyang-etal-2019-robust,Zaki2020AmharicAT,uthus-etal-2023-mlongt5,hasan-etal-2021-xl,pfeiffer-etal-2023-mmt5}, and spelling or grammatical error correction~\citep{adouane-etal-2019-normalising,Gezmu2021ManuallyAS,Keita2024GrammaticalEC}—are among the least explored for African languages. Within the text generation category, existing work also includes data-to-text~\citep{gehrmann-etal-2023-tata}, question generation~\citep{asai-etal-2024-buffet}, title or topic generation~\citep{adebara-etal-2024-cheetah,meyer-etal-2024-nglueni}, paragraph generation, and automatic diacritization~\citep{Olawole2024YADLT,Ojo2023AfroBenchHG}. 

\textbf{Others}: There is also work on discourse relation classification for Nigerian Pidgin~\citep{saeed2024implicitdiscourserelationclassification}, and in the context of computational social science and cultural analytics on historical text analysis~\citep{Schoots2023AnalyzingPF}, dialectal variation analysis~\citep{gugliotta-dinarelli-2020-tarc-incrementally,Eisenstein2023MD3TM}, and modeling social linguistic factors in creoles~\citep{ndubuisi-obi-etal-2019-wetin}.

\section{Other low-resource techniques}
\label{app:lowrestec}

\textbf{Meta Learning} has increasingly been applied to African and other low-resource languages, addressing challenges like scarce labeled data~\citep{lux24_interspeech}, noisy supervision~\citep{zhu-etal-2023-meta}, and cross-lingual transfer~\citep{wu-etal-2023-good}. Approaches range from cross-lingual adaptation in NLP tasks (e.g., text classification~\citep{zhu-etal-2023-meta}, question answering~\citep{wu-etal-2023-good}, dependency parsing~\citep{choenni-etal-2023-cross}, text generation~\citep{maurya-desarkar-2022-meta}) to low-resource speech recognition~\citep{CHEN2024101576,CHEN2024101648} and even text-to-speech synthesis~\citep{lux24_interspeech}, using techniques such as MAML, task-based meta-loss optimization, adversarial representation learning, and dynamic subnetworks. These methods enable models to generalize from resource-rich languages to underrepresented African languages, improve performance in zero- or few-shot settings, and handle noisy or weak supervision effectively.

\textbf{Annotation Projection (AP)} a technique used to transfer labels from a labeled source dataset to an unlabeled target dataset is also used for African  languages. Text classification benchmarks covering several African languages like SIB-200~\citep{adelani-etal-2024-sib} and Taxi-1500~\citep{ma2023taxi1500} are based on this technique, and it has also been shown to work for sequence labeling~\citep{garcia-ferrero-etal-2023-projection}.

\textbf{Ensemble Learning} a technique that combines multiple models usually either by aggregating predictions or representations—is also used in some of the surveyed literature, including approaches like model fusion~\citep{rathore-etal-2023-zgul,tran-etal-2021-facebook}.

\textbf{Multi-task learning (MTL)} an approach where a model is trained to perform multiple tasks simultaneously, leveraging shared knowledge across related tasks is also in the reviewed papers~\citep{Dossou2023FonMTLTM,Adebara2023ImprovingAL,Aduragba2023ImprovingHM,chen-etal-2024-interplay,wang-etal-2024-afrimte}.

\textbf{Active Learning} a machine learning approach where models dynamically select the data to train-was shown to be effective for African languages in tasks such as morphophonological processing~\citep{mirbostani-etal-2023-deep}, NER~\citep{Kholodna2024LLMsIT}, language modeling~\citep{dossou-etal-2022-afrolm}, and MT~\citep{chimoto-bassett-2022-comet}.

\textbf{Pruning \& Compression} \citep{Awobade2024WhatHW} studied the effect of compression of language, \citep{Mohammadshahi-etal-2022-compressed} studied the impact of NMT model compression on translation quality, \citep{Ahia-etal-2021-low-resource}

\textbf{Dictionary-based NLP} There are a few papers in this category. Some focus on creating lexicons for African languages~\citep{emezue-etal-2024-igboapi}, while others explore how to use these existing lexicons—particularly for tasks such as data augmentation~\citep{reid-etal-2021-afromt, wang-etal-2022-expanding}. Lexicon-based augmentation has proven useful for generating more data for language modeling. Additionally, a few works have shown their usefulness in tasks such as sentiment and emotional analysis~\citep{teodorescu-mohammad-2023-evaluating,mabokela-etal-2023-investigating} for African languages. 

\textbf{Rule-based NLP} Several papers involve rule-based NLP in different contexts, including grammar induction~\citep{herrera-etal-2024-sparse} and development~\citep{bamutura-etal-2020-towards} through corpus-based rule extraction, rule-based syntactic parsing using formal grammars~\citep{dione-2020-implementation}, and event extraction systems that combine handcrafted rules with machine learning~\citep{tadesse-etal-2020-event}. Additionally, rule-based approaches are evaluated for grammatical error correction. These studies showcase the diverse roles rule-based methods continue to play across linguistic resource creation and tasks.

\begin{figure}[t]
\setlength{\belowcaptionskip}{-10pt}
    \centering   \includegraphics[scale=0.35]{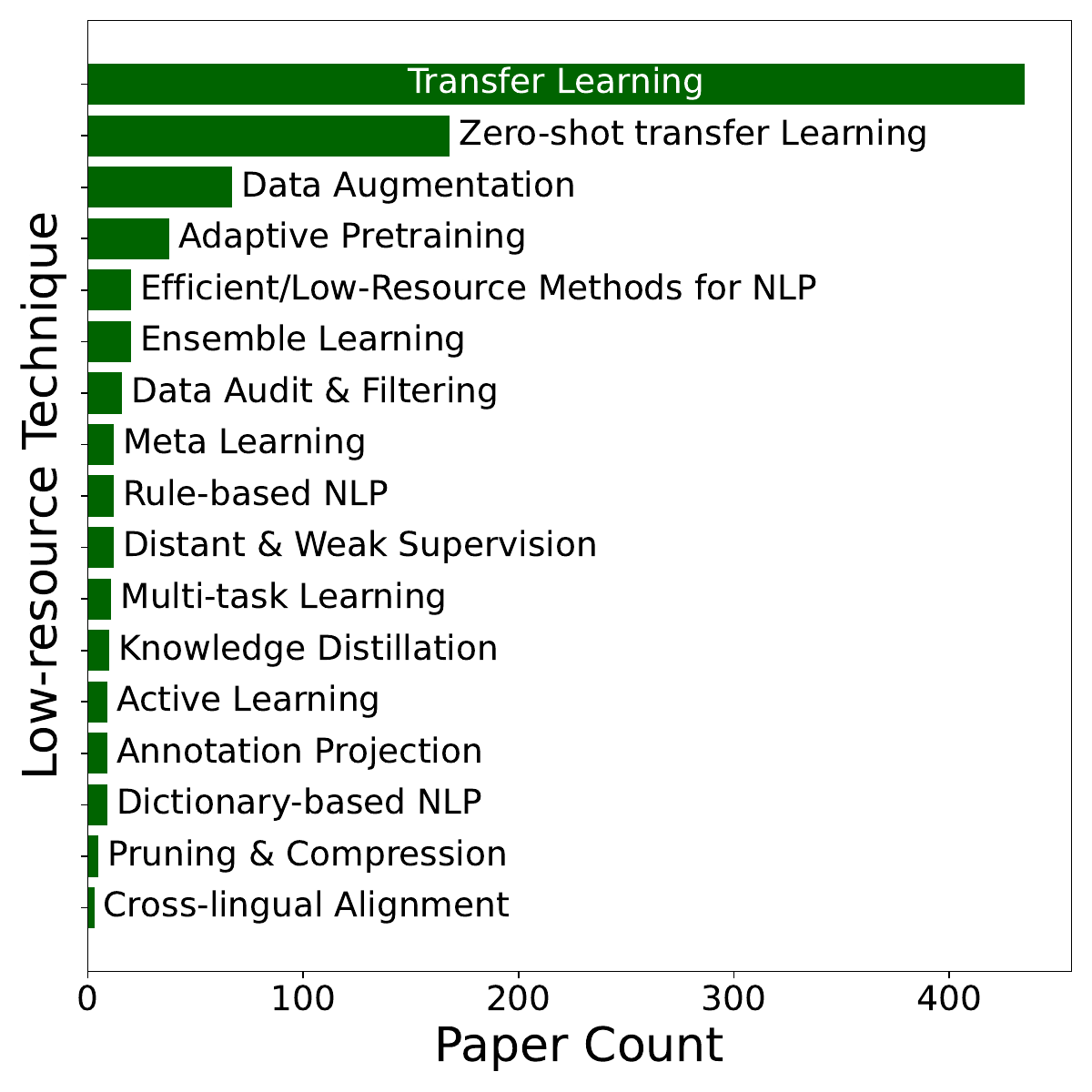}
    \caption{Low-resource techniques used in the analyzed papers.\looseness-1}
    \label{fig:lowrestech}
\end{figure}

\section{Comprehensive list of available African language resources}
\label{app:datalist}
Several language resources have been created for different languages over the last half decade. These resources include datasets, models, toolkits, and lexicons, all designed for various tasks. Some of these developments are supported by different funding sources and community initiatives such as HausaNLP\footnote{\url{https://hausanlp.github.io/}}, and GhanaNLP\footnote{\url{https://ghananlp.org/}}.

\subsection{Comprehensive list of datasets and benchmarks}
In \Cref{tab:app_resource1}, we provide a list of some of the created and publicly released labeled and unlabeled text and audio datasets for African languages across various tasks. While some of these datasets are part of large, curated multilingual language resources, others are tailored to a single language or a small group of languages. The list also includes datasets created for African variants of English~\citep{Afonja2021LearningNA,Eisenstein2023MD3TM,Ogun20241000AV}. Although a few works were done on African French~\citep{Alume2023ExploringTI,Maison2023ImprovingAS} and Portuguese~\citep{Hagemeijer2022ThePC}, there were no datasets created. 

Although they are not highlighted in the tables, there are also multimodal, multilingual, and multitask benchmarks that have been created over the years surveyed—especially in the era of LLMs—to evaluate their generalization and reasoning abilities. They include benchmarks such as BLEnD~\citep{Myung2024BLEnDAB}, XTREME~\citep{Hu2020XTREMEAM}, XTREME-UP~\citep{ruder-etal-2023-xtreme}, IrokoBench~\citep{adelani2025irokobenchnewbenchmarkafrican}, MEGAVERSE~\citep{ahuja-etal-2024-megaverse}, and AfroBench~\citep{Ojo2023AfroBenchHG}. There are those specifically tailored towards dialects of languages includes dialects of African languages~\citep{Mousi2024AraDiCEBF,faisal-etal-2024-dialectbench} and also creoles which are spoken across Africa~\citep{robinson-etal-2024-kreyol,lent-etal-2024-creoleval}. 

In addition to data collection, there are also articles focused on African language data curation methodology, identifying challenges and proposing solutions such as creating platforms to support creation and storage~\citep{Tchiaze2020BuildingCR,Griesel2020NavigatingCO,Griscom2020MobilizingMO,Marivate2020InvestigatingAA}.

\subsection{Comprehensive list of models, toolkits and platforms}
\label{sec:appmodels}
Several models have been developed and published for African languages in the surveyed period, spanning a variety of NLP tasks. Two key areas that have seen a proliferation of models are machine translation and language modeling. These models include both language-specific~\citep{emezue-dossou-2020-ffr,adelani-etal-2022-thousand} and multilingual systems~\citep{Fan2020BeyondEM,elmadany-etal-2024-toucan,mohammadshahi-etal-2022-small}, largely enabled by the creation of extensive multilingual datasets—such as web-mined bi-texts.

Although these multilingual resources can be used directly, multiple studies have shown that such models tend to underperform in African languages compared to high-resource languages such as German, French and English~\citep{adelani-etal-2024-sib,adelani2025irokobenchnewbenchmarkafrican}. To address this disparity, transfer learning has emerged as an effective strategy: pretrained multilingual models are adapted to low-resource African languages via task-specific fine-tuning or adaptive pretraining, followed by fine-tuning. These transfer learning techniques help bridge the performance gap by allowing models to leverage knowledge from high-resource languages or related tasks.

Examples of pretrained translation models used in this way include M2M-100~\citep{Fan2020BeyondEM} and NLLB~\citep{nllb2024scaling}, which support a wide range of African languages. For language modeling, models vary across architectural paradigms—encoder-only~\citep{ogueji-etal-2021-small,oladipo2024scaling,conneau-etal-2020-unsupervised,imanigooghari-etal-2023-glot500}, decoder-only~\citep{xue-etal-2021-mt5,adebara-etal-2024-cheetah}, and encoder-decoder—and in their scope as monolingual~\citep{  Abdaoui2021DziriBERTAP,chukwuneke-etal-2022-igbobert,martin-etal-2022-swahbert,nzeyimana-niyongabo-rubungo-2022-kinyabert,Olawole2024YADLT} or multilingual systems~\citep{ogueji-etal-2021-small,oladipo2024scaling,adebara-etal-2023-serengeti}.

In recent years, including the period covered by this survey, there has been a surge in the development of general-purpose LLMs, often multilingual and containing billions of parameters. These models support zero-shot and few-shot learning through the in-context learning paradigm, making them adaptable to a wide range of downstream tasks with minimal task-specific supervision.

These LLMs include open-source and proprietary models. Although the training data and language coverage for proprietary models are often undisclosed, open models are typically trained on data from high-resource languages. Evaluations of both open and closed models on a range of tasks in African languages consistently show relatively poor performance, highlighting persistent challenges in cross-lingual generalization and data representation for low-resource languages.

We highlight several models' names in \Cref{{tab:app_resource2}}. For models and datasets covering a single language, we include the ISO-3 code for that language. In cases where the data pertain to dialects of non-indigenous African languages, such as African Portuguese (APs) and African English (AEAs), we use appropriate dialectal labels instead.

\begin{table}[ht]
\small\centering
\scalebox{0.6}{
  \begin{tabular}{lll}
  \toprule
  \textbf{Name} & \textbf{Task}  & \textbf{Number}  \\
  \midrule
  \multicolumn{3}{l}{\emph{\textbf{NLU Tasks}}} \\
  Taxi-1500~\citep{ma2023taxi1500} & text classification & 420 \\
  SIB-200~\citep{adelani-etal-2024-sib} & text classification & 53 \\
  MasakhaNEWS~\citep{adelani-etal-2023-masakhanews} & topic classification & 16 \\
  KINNEWS and KIRNEWS~\citep{niyongabo-etal-2020-kinnews} & text classification & 2 \\
  Amharic TC~\citep{Azime2021AnAN} & text classification & 1 (amh) \\
Emakhuwa TC~\citep{ali-etal-2024-building} & text classification & 1 (vmw) \\
  AfriNLI~\citep{adelani2025irokobenchnewbenchmarkafrican} & NLI & 16 \\
  AfriSenti~\citep{muhammad-etal-2023-afrisenti} & sentiment analysis & 14 \\
  NaijaSenti~\citep{muhammad-etal-2022-naijasenti} & sentiment analysis & 4 \\ 
  NollySenti~\citep{shode-etal-2023-nollysenti} & sentiment analysis & 4 \\
  OMCD~\citep{Sibeko2024DevelopingAT} & off. lang. detection & 1 (ary) \\
  \multicolumn{3}{l}{\emph{\textbf{MT Tasks}}} \\
  FLORES-200~\citep{nllb2024scaling} & MT & 53 \\ 
  NTREX-128~\citep{federmann-etal-2022-ntrex} & MT & 24 \\ 
  MAFAND-MT~\citep{adelani-etal-2022-thousand} & MT & 16 \\
  TICO-19~\citep{anastasopoulos-etal-2020-tico} & MT & 12 \\
  WikiMatrix & MT & \\
  MENYO-20k~\citep{akinade-etal-2023-varepsilon} & MT & 1 (yor) \\
  AGE~\citep{Ademtew2024AGEAG} & MT & 1 (amh) \\
  Emakhuwa News MT~\citep{ali-etal-2024-building} & MT & 1 (vmw) \\
  Emakhuwa-FLORES~\citep{ali-etal-2024-expanding} & MT & 1 (vmw) \\
  \multicolumn{3}{l}{\emph{\textbf{Question Answering}}} \\
  Belebele~\citep{bandarkar-etal-2024-belebele} & MRC & 53 \\
  AfriMMLU~\citep{adelani2025irokobenchnewbenchmarkafrican} & QA & 16 \\
  AfriQA~\citep{ogundepo-etal-2023-cross} & QA & 10 \\
  NaijaRC~\citep{aremu2024naijarc} & MRC & 3 \\
  TyDi QA~\citep{clark-etal-2020-tydi}  & QA & 1 (swh) \\
  KenSwQuAD~\citep{Wanjawa2023kensquad}  & QA & 1 (swh) \\
  Amh-QuAD~\citep{taffa-etal-2024-low}   & QA & 1 (amh) \\
  {Y}-{NQ}~\citep{costa-jussa-etal-2025-y} & QA & 1 (yor) \\
  \multicolumn{3}{l}{\emph{\textbf{Reasoning}}} \\
  AfriMGSM~\citep{adelani2025irokobenchnewbenchmarkafrican} & Math & 17 \\
  LINGOLY~\citep{beanLINGOLYBenchmarkOlympiadLevel2024} & Linguistic & 13 \\
  MGSM~\citep{shi2023language} & Math & 1 (swh) \\
  XCOPA~\citep{beanLINGOLYBenchmarkOlympiadLevel2024} & Commonsense & 1 (swh) \\
  \multicolumn{3}{l}{\emph{\textbf{Information Extraction}}} \\
  MasakhaNER 2.0~\citep{adelani-etal-2022-masakhaner} & IE & 20 \\
  MasakhaNER~\citep{adelani-etal-2021-masakhaner} & IE & 10 \\
  MphayaNER~\citep{Mbuvha2023MphayaNERNE} & NER & 1 (ven) \\
  \multicolumn{3}{l}{\emph{\textbf{Information Retrieval}}} \\
  AfriQA~\citep{ogundepo-etal-2023-cross} & IR & 10 \\
  CIRAL~\citep{adeyemi2024ciral} & IR & 4 \\
  MIRACL~\citep{zhang-etal-2023-miracl} & IR & 2 \\
  Mr. TyDi~\citep{zhang-etal-2021-mr} & IR & 1 (swh) \\
  \multicolumn{3}{l}{\emph{\textbf{Syntax: Parsing}}} \\
  MasakhaPOS~\citep{dione-etal-2023-masakhapos} & POS & 20 \\
  NguniPOS~\citep{Sibeko2024DevelopingAT} & POS & 4 \\
  YTB~\citep{ishola-zeman-2020-yoruba} & DP & 1 (yor) \\
  Tswana TB~\citep{gaustad-etal-2024-first} & DP & 1 (tsn) \\
  Swahili TB~\citep{steimel-etal-2023-towards} & DP & 1 (swh) \\
  CACO~\citep{Gezmu2021ContemporaryAC} & POS & 1 (amh) \\
  PALMA~\citep{hagemeijer-etal-2022-palma} & POS & 0 (APs) \\
  NArabizi treebank~\citep{seddah-etal-2020-building} & DP & 0 (NAA) \\ 
  \multicolumn{3}{l}{\emph{\textbf{\{Spoken\} Language/Dialect Identification}}} \\
  OpenLID~\citep{burchell-etal-2023-open} & LID & 53 \\
  VoxLingua107~\citep{Valk2020VOXLINGUA107AD} & SLID & 13 \\

  \multicolumn{3}{l}{\emph{\textbf{Text Generation}}} \\
  
  AfriWOZ~\citep{Adewumi2023AfriWOZCF} & Dialogue Generation & 6 \\
  TATA~\citep{gehrmann-etal-2023-tata} & Table-to-text & 6 \\
  YAD~\citep{Olawole2024YADLT} & ADR & 1 (yor) \\
  SepediRadio Corpus~\citep{Ramalepe2023TheAO} & Text Generation & 1 (nso) \\

  \multicolumn{3}{l}{\emph{\textbf{Text Summarization}}} \\
  CrossSum~\citep{bhattacharjee-etal-2023-crosssum} & Summarization & 6 \\
  XL-Sum~\citep{hasan-etal-2021-xl} & Summarization & 6 \\
  
  \multicolumn{3}{l}{\emph{\textbf{Speech Tasks}}} \\ 
  BibleMMS~\citep{lux24_interspeech} & TTS & 59 \\
  Yodas~\citep{Li2023YodasYD} & ASR & 56 \\
  FLEURS~\citep{Conneau2022FLEURSFL} & ASR & 20 \\
  FLEURS-R~\citep{ma24c_interspeech} & TTS & 20 \\
  SpeechTaxi~\citep{Keller2024SpeechTaxiOM} & Speech Classification & 8 \\
  SD-QA~\citep{faisal-etal-2021-sd-qa} & Spoken QA & 7 (AEAs) \\
  BibleTTS~\citep{meyer22c_interspeech} & TTS & 6 \\
  Zambezi Voice~\citep{sikasote23_interspeech} & ASR & 4 \\
  African Voices~\citep{ogayo22_interspeech} & TTS & 6 \\
  Nicolingua~\citep{doumbouya2021usingradio} & ASR & 3 \\
  Kallaama~\citep{gauthier-etal-2024-kallaama} & ASR & 3 \\
  LRSC~\citep{Kimanuka2023SpeechRD} & ASR & 1 (lin) \\
  VoxMg~\citep{Ramanantsoa2023VoxMgAA} & ASR & 1 (plt) \\
  EYASE~\citep{AbdelHamid2020EgyptianAS} & Sp. emotion recog & 1 (arz) \\
  SADA~\citep{Alharbi2024SADASA} & ASR & 1 (arz) \\
  AfriSpeech-200~\citep{olatunji-etal-2023-afrispeech} & ASR & 0 (AEAs) \\
  EdAcc~\citep{Sanabria2023TheEI} & ASR & 0 (AEAs) \\
  Afro-TTS~\citep{ogun24_interspeech} & TTS & 0 (AEAs) \\  
  PidginASR~\citep{Ajisafe2020TowardsET}  & ASR & 1 (pcm) \\ 
 \multicolumn{3}{l}{\emph{\textbf{Unlabeled Text / Language Modeling}}} \\    
  GlotCC~\citep{Kargaran2024GlotCCAO} & LM & 251 \\
  EMMA-500~\citep{Ji2024EMMA500EM} & LM & 217 \\
  FineWeb~\citep{Penedo2024TheFD}  & LM & 90 \\
  MADLAD-400~\citep{Kudugunta2023MADLAD400AM} & LM & 87 \\
  Bloom~\citep{leong-etal-2022-bloom} & LM & 87 \\
  WURA~\citep{oladipo-etal-2023-better}  & LM & 16 \\
  mC4~\citep{xue-etal-2021-mt5} & LM & 13 \\
  CulturaX~\citep{nguyen-etal-2024-culturax} & LM & 7 \\
  CulturaY~\citep{nguyen2024culturay} & LM & 3 \\
  HPLT~\citep{de-gibert-etal-2024-new} & LM & 3 \\
  TLMD~\citep{gaim_2021_5139094} & LM & 1 (tir) \\
 \multicolumn{3}{l}{\emph{\textbf{Unlabeled Speech / Speech Representation Learning}}} \\   
 MMS ulab v2~\citep{pratap2023scalingspeechtechnology1000,chen-etal-2024-towards-robust} & SRL & 1450 \\
 Jesus Dramas~\citep{chen-etal-2024-towards-robust} & SRL & 88 \\
 Nicolingua~\citep{doumbouya2021usingradio} & SRL & 10 \\
 Zambezi Voice~\citep{sikasote23_interspeech} & SRL & 4 \\
 CSRC~\citep{Kimanuka2023SpeechRD} & SRL & 4 \\
  \bottomrule
  \end{tabular}
  }
  \vspace{-2mm}
  \caption{Some publicly available datasets including African languages published between 2019 and 2024.}
  \vspace{-5mm}
  \label{tab:app_resource1}
\end{table}

\begin{table}[ht]
\small\centering
\scalebox{0.65}{
  \begin{tabular}{lll}
  \toprule
  \textbf{Name} & \textbf{Task}  & \textbf{Number}  \\
  \midrule
  \multicolumn{3}{l}{\emph{\textbf{Translation Models}}} \\
  Toucan~\citep{elmadany-etal-2024-toucan} & MT & 517 \\
  NLLB-200~\citep{nllb2024scaling} & MT & 53  \\
  MADLAD-MT~\citep{Kudugunta2023MADLAD400AM} & MT & 19 \\
  M2M-100~\citep{Fan2020BeyondEM} & MT & 17 \\
  SMaLL-100~\citep{mohammadshahi-etal-2022-small} & MT & 17 \\

  \addlinespace
  \multicolumn{3}{l}{\emph{\textbf{Encoder Language Models}}} \\
  Serengeti~\citep{adebara-etal-2023-serengeti} & Encoder LM & 517 \\
  Glot500-m~\citep{imanigooghari-etal-2023-glot500} & Encoder LM & 517 \\
  AfroXLMR-76L~\citep{adelani-etal-2024-sib} & Encoder LM & 76 \\
  AfroLM~\citep{dossou-etal-2022-afrolm} & Encoder LM & 23 \\
  AfriBERTaV2~\citep{oladipo2024scaling} & Encoder LM & 23 \\
  AfroXLMR~\citep{alabi-etal-2022-adapting} & Encoder LM & 17 \\
  AfriBERTa~\citep{ogueji-etal-2021-small} & Encoder LM & 11 \\
  RemBERT~\citep{Chung2020RethinkingEC} & Encoder LM & 11 \\
  mDeBERTaV3~\citep{He2021DeBERTaV3ID} & debrt LM & 11 \\
  XLM-R~\citep{conneau-etal-2020-unsupervised} & Encoder LM & 8 \\
  EthioLLM~\citep{tonja-etal-2024-ethiollm}  & Encoder LM & 5 \\
  Nguni-XLMR~\citep{meyer-etal-2024-nglueni}  & Encoder LM & 4 \\
  KinyaBERT~\citep{nzeyimana-niyongabo-rubungo-2022-kinyabert} & Encoder LM & 1 (kin) \\
  PuoBERTa~\citep{Marivate2023PuoBERTaTA} & Encoder LM & 1 (tsn) \\
  SwahBERT~\citep{martin-etal-2022-swahbert} & Encoder LM & 1 (swh) \\
  IgboBERT~\citep{chukwuneke-etal-2022-igbobert} & Encoder LM & 1 (ibo) \\
  DziriBERT~\citep{Abdaoui2021DziriBERTAP} & Encoder LM & 1 (arq) \\
  EgyBERT~\citep{Qarah2024EgyBERTAL} & Encoder LM & 1 (arz) \\

  \addlinespace
  \multicolumn{3}{l}{\emph{\textbf{Decoder Language Models}}} \\
  MADLAD~\citep{Kudugunta2023MADLAD400AM} & Decoder LM & 87 \\
  Goldfish~\citep{Chang2024GoldfishML} & Decoder LM & 63 \\
  BLOOM~\citep{Scao2022BLOOMA1} & Decoder LM & 22 \\
  XGLM~\citep{lin-etal-2022-shot} & Decoder LM & 20 \\
  LLaMAX~\citep{lu-etal-2024-llamax} & Decoder LM & 20 \\
  LLaMAX-Alpaca~\citep{lu-etal-2024-llamax} & Decoder LM/MT & 20 \\
  {A}fri{I}nstruct~\citep{uemura-etal-2024-afriinstruct} & Decoder LM & 19 \\
  AfroLLaMa\footnote{\url{https://huggingface.co/Jacaranda/AfroLlama_V1}} & Decoder LM & 5 \\
  InkubaLM~\citep{Tonja2024InkubaLMAS} & Decoder LM & 5 \\
  mGPT~\citep{shliazhko-etal-2024-mgpt} & Decoder LM & 3 \\
  Walia-LLM~\citep{azime-etal-2024-walia} & Decoder LM & 1 (amh) \\ 
  Amharic-LLaMA~\citep{Andersland2024AmharicLA} & Decoder LM & 1 (amh) \\

\addlinespace
  \multicolumn{3}{l}{\emph{\textbf{Encoder-Decoder Language Models}}} \\
  Cheetah~\citep{adebara-etal-2024-cheetah} & Enc-Dec LM & 517 \\
  AfriMBART~\citep{adelani-etal-2022-thousand} & Enc-Dec LM & 17 \\
  Afri\{M,By\}T5~\citep{adelani-etal-2022-thousand} & Enc-Dec LM & 17 \\
  AfriTeVa V2~\citep{oladipo-etal-2023-better} & Enc-Dec LM & 16 \\
  Aya-101~\citep{ustun-etal-2024-aya} & Enc-Dec LM & 15 \\
  mT0~\cite{muennighoff-etal-2023-crosslingual} & Enc-Dec LM & 14 \\
  mT5~\citep{xue-etal-2021-mt5} & Enc-Dec LM & 13 \\
  ByT5~\citep{xue-etal-2022-byt5} & Enc-Dec LM & 13 \\
  mLongT5~\cite{uthus-etal-2023-mlongt5} & Enc-Dec LM & 13 \\
  AfriTeVa~\citep{jude-ogundepo-etal-2022-afriteva} & Enc-Dec LM & 11 \\
  EthioMT5~\citep{tonja-etal-2024-ethiollm}  & Enc-Dec LM & 5 \\
  OyoT5~\citep{Olawole2024YADLT}  & Enc-Dec LM & 1 (yor) \\

\addlinespace
  \multicolumn{3}{l}{\emph{\textbf{Speech encoders}}} \\
  XEUS~\citep{chen-etal-2024-towards-robust} & Speech Rep. & 1439 \\
  MMS~\citep{pratap2023scalingspeechtechnology1000} & Speech Rep. & 1396 \\
  AfriHuBERT~\citep{Alabi2024AfriHuBERTAS} & Speech Rep. & 1226 \\
  SSA-HuBERT~\citep{caubriere2024ssaspeechssl} & Speech Rep. & 21 \\
  mHuBERT-147~\citep{boito2024mhubert} & Speech Rep. & 16 \\
  XLSR-128~\citep{babu22_interspeech} & Speech Rep. & 12 \\
  \addlinespace
  \midrule
  \multicolumn{3}{l}{\emph{\textbf{Toolkits}}} \\
  OccGen~\citep{NEURIPS2022_09933f07} & Bias, Data selection & 1 (swh) \\
  Lesan~\citep{Hadgu2021LesanM} & MT & 1 (amh) \\
  AfroLID~\citep{adebara-etal-2022-afrolid} & LID & 517 \\
  OpenLID~\citep{burchell-etal-2023-open} & LID & 53 \\
  GlotScript~\citep{kargaran-etal-2024-glotscript-resource} & LID & 2250 \\
  \addlinespace
  \multicolumn{3}{l}{\emph{\textbf{Platforms}}} \\
  Mozilla Common Voice~\citep{ardila-etal-2020-common} & - & - \\
  Lanfrica~\citep{Emezue2020LanfricaAP} & - & - \\
  \bottomrule
  \end{tabular}
  }
  \vspace{-2mm}
  \caption{Some publicly available models, toolkits and platforms including African languages published between 2019 and 2024.}
  \vspace{-5mm}
  \label{tab:app_resource2}
\end{table}

\end{document}